# Research Highlights

- A PSO and pattern search based memetic algorithm is proposed

- Probabilistic selection strategy is proposed to select individuals for local search

- Local refinement with PS and probabilistic selection strategy are confirmed

- Experimental results show that the proposed MA outperforms established counterparts

# A PSO and Pattern Search based Memetic Algorithm for SVMs Parameters Optimization

Yukun Bao[*], Zhongyi Hu, Tao Xiong

Department of Management Science and Information Systems,

School of Management, Huazhong University of Science and Technology, Wuhan,

P.R.China, 430074

**Abstract**

Addressing the issue of SVMs parameters optimization, this study proposes an efficient memetic algorithm based on Particle Swarm Optimization algorithm (PSO) and Pattern Search (PS). In the proposed memetic algorithm, PSO is responsible for exploration of the search space and the detection of the potential regions with optimum solutions, while pattern search (PS) is used to produce an effective exploitation on the potential regions obtained by PSO. Moreover, a novel probabilistic selection strategy is proposed to select the appropriate individuals among the current population to undergo local refinement, keeping a well balance between exploration and exploitation. Experimental results confirm that the local refinement with PS and our proposed selection strategy are effective, and finally demonstrate effectiveness and robustness of the proposed PSO-PS based MA for SVMs parameters optimization.

[*] Corresponding author: Tel: +86-27-62559765; fax: +86-27-87556437.

Email: yukunbao@mail.hust.edu.cn or y.bao@ieee.org

# 1. Introduction

Support vector machines (SVMs), first presented by Vapnik [1] based on Statistical Learning Theory (SLT) and Structural Risk Minimization principle (SRM), solve the classification problem by maximizing the margin between the separating hyper-plane and the data. SVMs implement the Structural Risk Minimization Principle (SRM), which seeks to minimize an upper bound of the generalization error by using penalty parameter *C* as trade-off between training error and model complexity. The use of kernel tricks enables the SVMs to have the ability of dealing with nonlinear features in high dimensional feature space. Due to the excellent generalization performance, SVMs have been widely used in various areas, such as pattern recognition, text categorization, fault diagnosis and so on. However, the generalization ability of SVMs highly depends on the adequate setting of parameters, such as penalty coefficient *C* and kernel parameters. Therefore, the selection of the optimal parameters is of critical importance to obtain a good performance in handling learning task with SVMs.

In this study, we mainly concentrate on the parameters optimization of SVMs, which has gained great attentions in the past several years. The most popular and universal method is grid search, which conducts an exhaustive search on the parameters space with the validation error (such as 5-fold cross validation error) minimized. Obviously, although it can be easily parallelized and seems safe [2], its computational cost scales exponentially with the number of parameters and the number of the sampling points for each parameter [3]. Besides, the performance of grid search is sensitive to the setting of the grid range and coarseness for each parameter, which are not easy to set without prior knowledge. Instead of minimizing the validation error, another streams of studies focused on minimizing the approximated bounds on generalization performance by numerical optimization methods [4-12]. The numerical optimization methods are generally more efficient than grid search for their fast convergence rate. However, this kind of methods can only be suitable for the cases that error bounds are differentiable and continuous with respect to the parameters in SVMs. It is also worth noting that the numerical optimization methods, such as gradient descent, may get stuck in local optima and highly depend on the starting points. Furthermore, experimental evidence showed that several established bounds methods could not compete with traditional 5-fold cross-validation method [6], which indicates inevitably gap between the approximation bounds and real error [13]. Recently, evolutionary algorithms such as Genetic algorithm (GA), particle swarm optimization (PSO), ant colony optimization (ACO) and simulated annealing algorithm (SA) have been employed to optimize the SVMs parameters for their better global search abilities against numerical optimization methods [14-19]. However, GAs and other evolutionary algorithms (EAs) are not guaranteed to find the global optimum solution to a problem, though they are generally good at finding "acceptable good" or near-optimal solutions to problems. Another drawback of EAs is that they are not well suited to perform finely tuned search, but on the other hand, they are good at exploring the solution space since they search from a set of designs and not from a single design.

Memetic Algorithms (MAs), first coined by Moscato [20,21], have been regarded as an promising framework that combines the Evolutionary Algorithms (EAs) with problem-specific local searcher (LS), where the latter is often referred to as a meme defined as a unit of cultural evolution that is capable of local refinements. From an optimization point of view, MAs are hybrid

EAs that combine global and local search by using an EA to perform exploration while the local search method performs exploitation. This has the ability to exploit the complementary advantages of EAs (generality, robustness, global search efficiency), and problem-specific local search (exploiting application-specific problem structure, rapid convergence toward local minima) [22]. Up to date, MAs have been recognized as a powerful algorithmic paradigm for evolutionary computing in a wide variety of areas [23-26]. In particular, the relative advantage of MAs over EAs is quite consistent on complex search spaces.

Since the parameters space of SVMs is often considered complex, it is of interest to justify the use of MAs for SVMs parameters optimization. There have been, if any, few works related to MAs reported in the literature of SVMs parameters optimization. In this study, by combining Particle Swarm Optimization algorithm (PSO) and Pattern Search (PS), an efficient PSO-PS based Memetic Algorithm (MA) is proposed to optimize the parameters of SVMs. In the proposed PSO-PS based MA, PSO is responsible for exploration of the search space and the detection of the potential regions with optimum solution, while a direct search algorithm, pattern search (PS) in this case, is used to produce an effective exploitation on the potential regions obtained by PSO. Furthermore, the problem of selecting promising individuals to experience local refinement is also addressed and thus a novel probabilistic selection strategy is proposed to keep a balance between exploration and exploitation. The performance of proposed PSO-PS based MA for parameters optimization in SVMs is justified on several benchmarks against selected established counterparts. Experimental results and comparisons demonstrate the effectiveness of the proposed PSO-PS based MA for parameters optimization of SVMs.

The rest of the study is organized as follows. Section 2 presents a brief review on SVMs. Section 3 elaborates on the PSO-PS based MA proposed in this study. The results with discussions are reported in Section 4. Finally, we conclude this study in Section 5.

**2. Support Vector Machines**

Consider a binary classification problem involving a set of training dataset $\{(\mathbf{x}_1, y_1), (\mathbf{x}_2, y_2)...(\mathbf{x}_n, y_n)\} \subset R^n \times R$, where $\mathbf{x}_i$ is input space, $y_i \in \{-1, 1\}$ is the labels of the input space $\mathbf{x}_i$ and $n$ denotes the number of the data items in the training set. Based on the structured risk minimization (SRM) principle [1], SVMs aim to generate an optimal hyper-plane to separate the two classes by minimizing the regularized training error:

$$\begin{aligned} \min \quad & \frac{1}{2}\|\mathbf{w}\|^2 + C\sum_{i=1}^{n}\xi_i \\ s.t. \quad & y_i(\langle \mathbf{w}, \mathbf{x}_i \rangle + b) \geq 1 - \xi_i, \quad i = 1, 2.....n \\ & \xi_i \geq 0, \quad i = 1, 2.....n \end{aligned} \quad (1)$$

Where, $\langle , \rangle$ denotes the inner product; $w$ is the weight vector, which controls the smoothness of the model; $b$ is a parameter of bias; $\xi_i$ is a non-negative slack variable which defines the permitted misclassification error. In the regularized training error given by Eq. (1), the first term, $\frac{1}{2}\|\mathbf{w}\|^2$, is the regularization term to be used as a measure of flatness or complexity of the function. The second term $\sum_{i=1}^{n}\xi_i$ is the empirical risk. Hence, $C$ is referred to as the penalty coefficient and it specifies the trade-off between the empirical risk and the regularization term.

According to Wolfe's Dual Theorem and the saddle-point condition, the dual optimization problem of the above primal one is obtained as the following quadratic programming form:

$$\begin{aligned}&\max\quad \sum_{i=1}^{n}\alpha_i - \frac{1}{2}\sum_{i,j=1}^{n}\alpha_i\alpha_j y_i y_j \langle \mathbf{x}_i, \mathbf{x}_j \rangle \\ &s.t.\quad \sum_{i=1}^{n}\alpha_i y_i = 0,\quad 0 \leq \alpha_i \leq C,\quad i=1,2.....n\end{aligned} \qquad (2)$$

Where $(\alpha_i)_{i \in n}$ are nonnegative Lagrangian multipliers that can be obtained by solving the convex quadratic programming problem stated above.

Finally, by solving Eq.(2) and using the trick of kernel function, the decision function can be defined as the following explicit form:

$$f(x) = \mathrm{sgn}\left(\sum_{i=1}^{n} y_i \alpha_i K(\mathbf{x}_i, \mathbf{x}) + b\right) \qquad (3)$$

Here, $K(\mathbf{x}_i, \mathbf{x})$ is defined as kernel function. The value of the kernel function is equivalent to the inner product of two vectors $\mathbf{x}_i$ an $\mathbf{x}_j$ in the high-dimensional feature space $\phi(\mathbf{x}_i)$ and $\phi(\mathbf{x}_j)$, that is, $K(\mathbf{x}_i, \mathbf{x}_j) = \langle \phi(\mathbf{x}_i), \phi(\mathbf{x}_j) \rangle$. The elegance of using the kernel function is that one can deal with feature spaces of arbitrary dimensionality without having to compute the map $\phi(\mathbf{x})$ explicitly. Any function that satisfies Mercer's condition [27] can be used as the kernel function. The typical examples of kernel function are as follows:

Linear kernel: $K(\mathbf{x}_i, \mathbf{x}_j) = \langle \mathbf{x}_i, \mathbf{x}_j \rangle$

Polynomial kernel: $K(\mathbf{x}_i, \mathbf{x}_j) = \left(\gamma \langle \mathbf{x}_i, \mathbf{x}_j \rangle + r\right)^d, \gamma > 0.$

Radial basis function (RBF) kernel: $K(\mathbf{x}_i, \mathbf{x}_j) = \exp\left(-\gamma \|\mathbf{x}_i - \mathbf{x}_j\|^2\right), \gamma > 0.$

Sigmoid kernel: $K(\mathbf{x}_i, \mathbf{x}_j) = \tanh\left(\gamma \langle \mathbf{x}_i, \mathbf{x}_j \rangle + r\right).$

Where, $\gamma$, $r$ and $d$ are kernel parameters. The kernel parameter should be carefully chosen as it implicitly defines the structure of the high dimensional feature space $\phi(\mathbf{x})$ and thus controls the complexity of the final solution [28]. Generally, among these kernel functions, RBF kernel is strongly recommended and widely used for its performance and complexity [2] and thus SVMs with RBF kernel function is the one studied in this study.

Overall, SVMs are a powerful classifier with strong theoretical foundations and excellent generalization performance. Note that before implementing the SVMs with RBF kernel, there are two parameters (penalty parameter $C$ and RBF kernel parameter $\gamma$) have to set. Previous studies show that these two parameters play an important role on the success of SVMs. In this study, addressing the selection of these two parameters, a PSO-PS based Memetic Algorithm (MA) is proposed and justified within the context of SVMs with RBF kernel function.

**3. Proposed Memetic Algorithms**

Memetic Algorithms (MAs), first coined by Moscato [20,21], have come to light as an union of population-based stochastic global search and local based cultural evolution, which are inspired by Darwinian principles of natural evolution and Dawkins notion of a meme. The meme is defined as a unit of cultural evolution that is capable of local/individual refinements. As designed as hybridization, MAs are expected to make full use of the balance between exploration and exploitation of the search space to complement the advantages of population based methods and local based methods. Nowadays, MAs have revealed their successes with high performance and superior robustness across a wide range of problem domains [29,30]. However, according to the No Free Lunch theory, the hybridization can be more complex and expensive to implement. Considering the simplicity in implementation of particle swarm optimizations (PSO) and its proven performance, being able to produce good solutions at a low computational cost [16,31-33], this study proposed a PSO based memetic algorithm with the pattern search (PS) as a local individual learner, to solve the parameters optimization problem in SVMs. The concept of the proposed Memetic algorithms is illustrated in Fig.1. In the proposed PSO-PS based MA, PSO is used for exploring the global search space of parameters, and pattern search is deserved to play the role of local exploitation based on its advantages of simplicity, flexibility and robustness.

In the following sub-sections, we will explain the implementation of the proposed PSO-PS MA for parameters optimization in details.

**3.1. Representation and Initialization**

In our MA, each particle/individual in the population is a potential solution to the SVMs parameters. And its status is characterized according to its position and velocity. The D-dimensional position for the particle $i$ at iteration $t$ can be represented as $\mathbf{x}_i^t =<x_{i1}^t,\ldots,x_{iD}^t>$. Likewise, the fly velocity (position change) for particle $i$ at iteration $t$ can be described as $\mathbf{v}_i^t =<v_{i1}^t,\ldots,v_{iD}^t>$. In this study, the position and velocity of one particle have only two dimensions that denote the two parameters ($C$ and $\gamma$) to be optimized in SVMs. A set of particles $\mathbf{P} =\{\mathbf{x}_1^t,\ldots,\mathbf{x}_M^t\}$ is called a swarm (or population).

Initially, the particles in the population are randomly generated in the solution space according to Eq.(4) and Eq.(5),

$$x_{id} = x_{\min,d} + r \times (x_{\max,d} - x_{\min,d}) \tag{4}$$

$$v_{id} = v_{\min,d} + r \times (v_{\max,d} - v_{\min,d}) \tag{5}$$

Where $x_{id}$ and $v_{id}$ are the position value and velocity in the $d$th dimension of the $i$th particle, respectively. $r$ is a random number in the range of [0,1]. $x_{\min,d}$ and $x_{\max,d}$ denotes the search space of the $d$th dimension (that is, the upper and lower limit of the parameters in SVMs), while $v_{\min,d}$ and $v_{\max,d}$ is used to constraint the velocity of the particle to avoid the particle flying outside the search space.

After the initialization of the population, the algorithm iteratively generates offspring using PSO based operator and undergoes local refinement with pattern search, which is to be discussed in the following sub-sections.

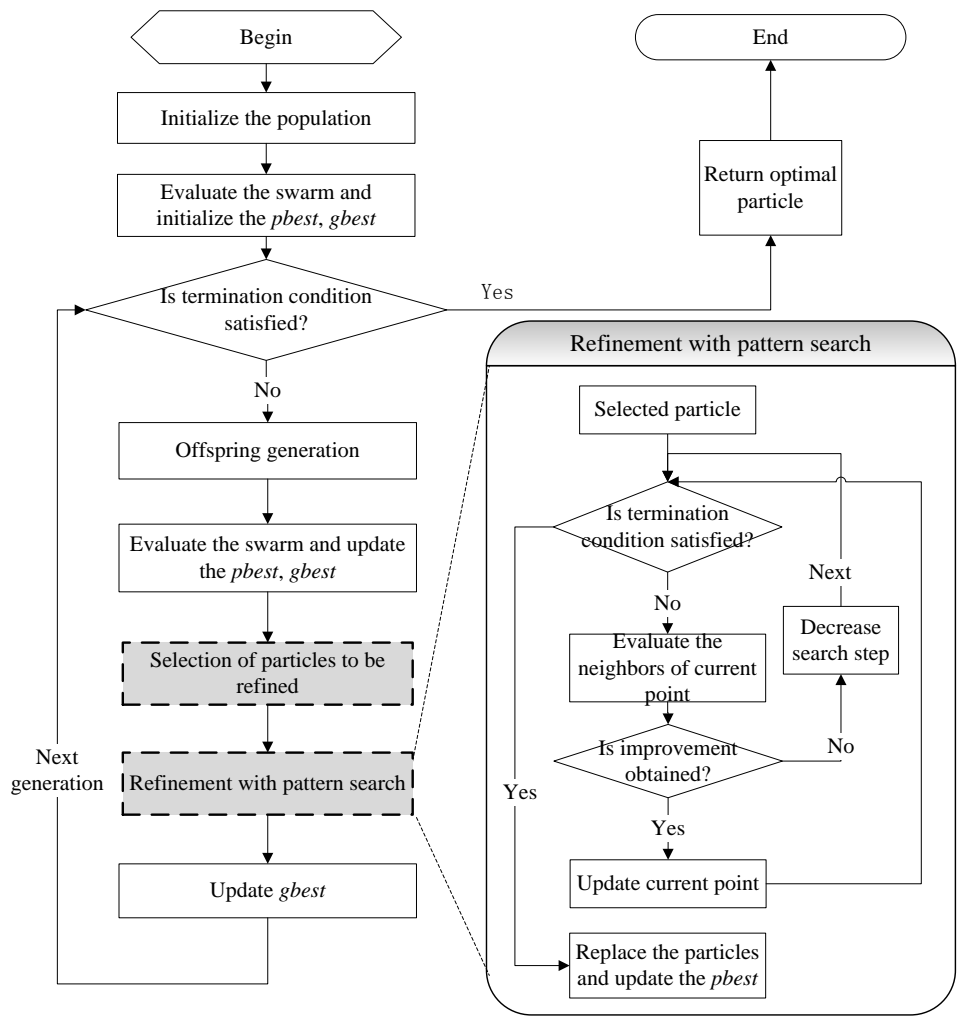

Fig.1. Flowchart of the proposed PSO-PS based Memetic Algorithm

## 3.2. Fitness Function

The objective of optimizing the parameters in a SVMs model is to maximize the generalization ability or minimize generalization error of the model. As cross validation can provide unbiased estimation of the generalization error, in this study, we take the k-fold cross validation error into consideration. Let *kCVMR* denotes the mean of the *k* folds' misclassification rate (*MR*). With the use of cross validation, *kCVMR* is deserved to be a criterion for generalization ability of a model. Hence, *kCVMR* is used as fitness function in this study. The fitness function and errors are shown as follows:

$$Fitness = kCVMR$$
$$kCVMR = \frac{1}{k}\sum_{j=1}^{k} MR_j \qquad (6)$$

Where, *k* is the size of folds. In this study, 5-fold cross validation is conducted, which is widely used and suggested in [34].

## 3.3. PSO based Operator

In the proposed PSO-PS based MA, PSO based operators are used to explore the search space. Particle Swarm Optimization (PSO) [35] is a population-based meta-heuristic that simulates social behavior such as birds flocking to a promising position to achieve precise objectives (e.g., finding food) in a multi-dimensional space by interacting among them.

To search for the optimal solution, each particle adjusts their flight trajectories by using the following updating equations:

$$v_{id}^{t+1} = w \times v_{id}^{t} + c_1 \times r_1 \times (p_{id} - x_{id}^{t}) + c_2 \times r_2 \times (p_{gd} - x_{id}^{t}) \qquad (7)$$

$$x_{id}^{t+1} = x_{id}^{t} + v_{id}^{t+1} \tag{8}$$

Where $c_1, c_2 \in \mathbb{R}$ are acceleration coefficients, $w$ is inertia weight, $r_1$ and $r_2$ are random numbers in the range of [0,1]. $v_{id}^{t}$ and $x_{id}^{t}$ denote the velocity and position of the $i$th particle in $d$th dimension at $t$th iteration, respectively. $p_{id}$ is the value in dimension $d$ of the best parameters combination (a particle) found so far by particle $i$. $\mathbf{p}_i = <p_{i1},\ldots,p_{iD}>$ is called personal best (*pbest*) position. $p_{gd}$ is the value in dimension $d$ of the best parameters combination (a particle) found so far in the swarm (**P**); $\mathbf{p}_g = <p_{g1},\ldots,p_{gD}>$ is considered as the global best (*gbest*) position. Note that each particle takes individual (*lbest*) and social (*gbest*) information into account for updating its velocity and position.

In the search space, particles track the individual's best values and the best global values. The process is terminated if the number of iteration reaches the pre-determined maximum number of iteration.

**3.4. Refinement with Pattern Search**

In the proposed PSO-PS based MA, pattern search is employed to conduct exploitation of the parameters solution space. Pattern search (PS) is a kind of numerical optimization methods that do not require the gradient of the problem to be optimized. It is a simple effective optimization technique suitable for optimizing a wide range of objective functions. It does not require derivatives but direct function evaluation, making the method suitable for parameters optimization of SVMs with the validation error minimized. The convergence to local minima for constrained problems as well as unconstrained problems has been proven in [36]. It investigates nearest neighbor parameter vectors of the current point, and tries to find a better move. If all neighbor points fail to produce a decrease, then the search step is reduced. This search stops until the search step gets sufficiently small, which ensuring the convergence to a local minimum. The pattern search is based on a pattern $P_k$ that defines the neighborhood of current points. A well often used pattern is five-point unit-size rood pattern which can be represented by the generating matrix $P_k$ in Eq.(9).

$$P_k = \begin{bmatrix} 1 & 0 & -1 & 0 & 0 \\ 0 & 1 & 0 & -1 & 0 \end{bmatrix} \quad (9)$$

The procedure of pattern search is outlined in Fig.2. $\Delta_0$ denotes the default search step of PS, $\Delta$ is a search step, $p_k$ is a column of $P_k$, $\Omega$ denotes the neighborhood of the center point. The termination conditions are: the maximum iteration is met, or the search step gets a predefined small value. To balance the amount of computational budget allocated for exploration versus exploitation, $\Delta_0/8$ is empirically selected as the minimum search step.

---

**Algorithm 1.** Refinement with pattern search for parameters optimization

---

**Begin**
    **Initialize**: predefine the default search step $\Delta_0$, initialize center point $p_0$;
    $\Delta = \Delta_0$ ;
    **While** (Termination conditions are not satisfied)
        $\Omega = \{p_0 + \Delta * p_k \mid \text{for each column } p_k \text{ in } P_k\}$ ;
        Evaluate the nearest neighbors in $\Omega$;
        **If** (there are improvements in the $\Omega$)
            Update the current center to the best neighbor in $\Omega$;    $\Delta = \Delta_0$ ;
        **Else**
            Decrease the search step $\Delta = \Delta_0/2$;
        **End If**
    **End While**
**End**

---

Fig.2. Pseudocode of pattern search for local refinement

## 3.5. A Probabilistic Selection Strategy

An important issue in MAs to be addressed is which particles from the population should be selected to undergo local improvement (see "*Selection of particles to be refined*" procedure in Fig.1). This selection directly defines the balance of evolution (exploration) and the individual learning (exploitation) under limited computational budget, which is of importance to the success of MAs. In previous studies, a common way is to apply a local search to all newly generated offspring, which suffers from much strong local exploitation. Another way is to apply local search to particles with a predefined probability $p_l$, which seems too blind and the $p_l$ is not easy to define [37]. Besides, as [38] suggests that those solutions that are in proximity of a promising basin of attraction received an extended budget, the randomly selected particles may needs more cost to reach the optimal. Hence, it seems a good choice that only the particles with the fitness larger than a threshold *v* are selected for local improvement [39], while the threshold is problem specific and some of the selected particles may be crowed in the same promising region.

To this end, a probabilistic selection strategy of selecting non-crowed individuals to undergo refinement is proposed in this study. Beginning with the best individual, each individual undergoes local search with the probability $p_l(x)$ (in Eq.(10)), a individual with better fitness has a more chance to be selected for exploitation. To avoid exploiting the same region simultaneously, once an individual is selected to be refined, those individuals crowd around the selected individual (i.e., the Euclidean distance is less than or equal to *r*) is eliminated from the population. The only parameter in the proposed strategy is *r*, which denotes the radius of the exploitation region. The *r* is selected empirically in this study. Detail of this strategy is illustrated in Fig.3.

Formally, the selection probability $p_l(x)$ of each solution $x$ in the current population **P** is specified by the following roulette-wheel selection scheme with the linear scaling:

$$p_l(x) = \frac{f_{\max}(\mathbf{P}) - f(x)}{\sum_{y \in \mathbf{P}}(f_{\max}(\mathbf{P}) - f(y))} \quad (10)$$

Where $f_{\max}(\mathbf{P})$ is the maximum (worse) fitness value among the current population **P**.

---

**Algorithm 2.** A Probabilistic Selection Strategy

---

**Input:** current population **P**, constant radius $r$

**Output:** a set of selected individuals that undergo local refinement $\Omega$;

**Begin**

    **Initialize**: **P'** as a copy of **P**; $\Omega$ as a null set ;

    **While** (**P'** is not NULL)

        Get the best fitness individual x belongs to **P'**, i.e., $x = \arg\min\{f(x) \mid x \in \mathbf{P'}\}$ ;

        Calculate $p_l(x)$ according to Eq.(10);

        **If** $p_l(x) \geq rand(0,1)$

            $\Omega = \Omega \cup \{x\}$ ;

            Remove $x$ from **P'**;

            **For** each individual $y$ in **P'**

                **If** $d(x, y) \leq r$, **then** remove $y$ from **P'**;

            **End For**

        **End If**

    **End While**

**End**

---

Fig.3. Pseudocode of proposed probabilistic selection strategy

## 4. Experimental Results and Discussions

### 4.1. Experimental Setup

In this study, we conduct two experiments to justify the proposed PSO-PS based MA as well as the proposed probabilistic selection strategy. The first experiment aims to investigate the performance of the four variants of the proposed PSO-PS based MA with different selection strategies and thus the best variant is determined. Then, the best selected variant is employed to compare against the established counterparts in literature in the second experiment.

LibSVM (Version 2.91) [34] is used for the implementation of SVMs. PSO and several variants of proposed MA are coded in MATLAB 2009b using computer with Intel Core 2 Duo CPU T5750, 2G RAM. As suggested by Hsu et al. [2], the search space of the parameters is defined as an exponentially growing space: $X = \log_2 C; Y = \log_2 \gamma$ and $-10 \leq X \leq 10; -10 \leq Y \leq 10$. Through initial experiments, the parameters in PSO and MA are set as follows. The acceleration coefficients $c_1$ and $c_2$ are both equals to 2. For $w$, a time varying inertia weight linearly decreasing from 1.2 to 0.8 is employed. The default search step ($\Delta$) in pattern search is set to 1. The radius of the exploitation region ($r$) used in the probabilistic selection strategy equals to 2.

Table 1 Details of the benchmark datasets

| Data sets | No. of attributes | No. of training set | No. of testing set | No. of groups |
|---|---|---|---|---|
| *Banana* | 2 | 400 | 4900 | 100 |
| *Breast cancer* | 9 | 200 | 77 | 100 |
| *Diabetis* | 8 | 468 | 300 | 100 |
| *Flare solar* | 9 | 666 | 400 | 100 |
| *German* | 20 | 700 | 300 | 100 |
| *Heart* | 13 | 170 | 100 | 100 |
| *Image* | 18 | 1300 | 1010 | 20 |
| *Ringnorm* | 20 | 400 | 7000 | 100 |
| *Splice* | 60 | 1000 | 2175 | 20 |
| *Thyroid* | 5 | 140 | 75 | 100 |
| *Titanic* | 3 | 150 | 2051 | 100 |
| *Twonorm* | 20 | 400 | 7000 | 100 |
| *Waveform* | 21 | 400 | 4600 | 100 |

To test the performance of the proposed PSO-PS based memetic algorithms, computational experiments are carried out against some well-studied benchmarks. In this study, thirteen commonly used benchmark datasets from IDA Benchmark Repository[1] are used. These datasets, having been preprocessed by [40], consist of 100 random groups (20 in the case of *image* and *splice* datasets) with training and testing sets (about 60%:40%) describing the binary classification problem. The dataset in each group has already been normalized to zero mean and unit standard deviation. For the dataset in each group, the training set is used to select the optimal parameters in SVM based on 5-fold cross validation and establish the classifier, and then the test set is used to assess the classifier with the optimal parameters. Table 1 summarizes the general information of these datasets.

**4.2. Experiment I**

In the proposed PSO-PS based MA, a pattern search is integrated into PSO. So it is worth to evaluate the influence of pattern search. Besides, in order to establish an efficient and robust MA, the selection of particles to be locally improved plays an important role. Hence, in the first experiment, a single PSO and four variants of proposed MA with different selection strategies are considered. The following abbreviations represent the algorithms considered in this sub-section.

*PSO*: a single PSO without local search;

*MA1*: all newly generated particles are refined with PS;

*MA2*: PS is applied to each new particle with a predefined probability $p_l$=0.1;

*MA3*: PS is only applied to the two best fitness particles;

*MA4*: PS is applied to the selected particles by probabilistic selection strategy proposed in section 3.5;

---

[1] http://www.raetschlab.org/Members/raetsch/benchmark/

*GS*: a grid search method on the search space, with the search step equals to 0.5.

This experiment is conducted on the first groups of the top six datasets (*Banana*, *Breast cancer*, *Diabetis*, *Flare solar*, *German* and *Heart*). The population size for PSO and each MA is fixed to 15. The stopping criterions of PSO are set as follows: the number of iterations reached 100 or there is no improvement in the fitness for 30 consecutive iterations. To compare the algorithms fairly, each MA stops when the number of the fitness evaluations reaches the maximum value that equals to 1500, or there is no improvement in the fitness value for 450 consecutive iterations. Besides, grid search (GS) evaluates the fitness 1681 times by sampling 41×41 points in the parameters' search space. For the purpose of reducing statistical errors, each benchmark is independently simulated 30 times, and the error rates and numbers of fitness evaluations are collected for comparison.

Table 2 describes the results obtained with the above algorithms. The column of 'error' shows the mean and standard deviation of the error rates (%) for 30 times. The "#eva" gives the mean and standard deviation of the numbers of fitness evaluations. The "**Ave**" row shows the average error rate (%) and average number of fitness evaluations on all datasets. The last row "**AveRank**" reports the average of the rank computed through the Friedman test for each algorithm on all datasets.

From Table 2, we can see that, no matter which kind of strategy is selected in MA, the performance in term of mean error rate is always better than that obtained by single PSO, which demonstrates the effectiveness of incorporating the local refinement with PS into PSO. Besides, from the view of standard deviation, it can be seen that the performance of proposed PSO-PS based MA is more stable than single PSO. As pattern search is employed to refine the individuals, the number of fitness evaluations of each MA is larger than that of the PSO, while it can be regarded as the cost to avoid premature convergence and conduct finely local refinement to obtain better performances. Actually, the average number of fitness evaluation of MA is larger than that of PSO by only at most 13% (e.g. *MA3*). Finally, compared with GS, both PSO and MA achieve better results than GS in terms of test error rates and number of fitness evaluations. The facts above imply that among the six algorithms investigated, four variants of proposed MA tend to achieve the smallest error rates with a little more fitness evaluations than that of PSO but still much less than that of grid search.

By comparing four variants of proposed MA, several results can be observed. It is interesting to note that *MA1*, where PS is applied to all newly generated individuals, does not achieve as good results as expected. This may be due to the premature convergence of the solutions because of the strong local refinement is applied to the population. *MA2* performs comparably with the *MA1*, which indicates that the randomly selected strategy by probability $p_l$ is also failed. *MA3* gains better results than *MA1* and *MA2*, and even has competitive results with *MA4* on some datasets (e.g. *Diabetis* dataset). Finally, *MA4* always performs the best in terms of the error rate on each dataset and the average error rate on all datasets, while the number of fitness evaluations does not increase too much than that of the other three variants of MA. All the above results suggest the success of our proposed probabilistic selection strategy used in *MA4*, highlighted by addressing the balance of exploration and exploitation in the proposed MA. Hence, we can conclude that the *MA4* with the proposed probabilistic selection strategy is effective and robust for solving parameters optimization in SVMs modeling.

In addition, the average computational time of each method is given in Table 3. It is obvious that all of the four variants of proposed MA consume more time than the single PSO in almost all the datasets, which is mainly due to refinement with pattern search. Among the four variants of the proposed MA with different selection strategy, the *MA4* consumes a little more time than the other three variants. Since the accuracy of a SVM model is often more important and the consumed time of *MA4* is still much less than that of the grid search, *MA4* is selected as the best one to perform a further comparison in subsection 4.3 with the established counterparts in the literature.

**4.3. Experiment II**

Furthermore, following the same experimental settings as in [12,13,40], the second experiment will directly compare the performance of the best variant as *MA4* of the proposed PSO-PS based MA with established algorithms in previous studies.

The experiment is conducted as follows. At first, on each of the first five groups of every benchmark dataset, the parameters are optimized using the proposed *MA4*, and then the final parameters are computed as the median of obtained five estimations. At last, the mean error rate and standard deviation of SVMs with such parameters on all 100 (20 for *image* and *splice* datasets) groups of each dataset are taken as the results of the corresponding algorithm. Table 4 and Fig.4 show the comparison of mean testing error rates (%) and the standard deviations (%) obtained by the proposed PSO-PS based MA and other algorithms proposed in previous studies [12,13,40]. For convenience of comparison, the best errors for every dataset are bolded. The first column reports the results obtained by two stages grid search based on 5 fold cross validation from [40]. The results of second and third column, both reported in [13], are respectively obtained by a radius-margin method and a fast-efficient strategy. "L1-CB" and "L2-CB" respectively show the results of hybridization of CLPSO and BFGS by minimizing the L1-SVMs and L2-SVMs generalization errors in [12]. Among them, although no method can outperform all the others on each dataset, the proposed PSO-PS based MA has lowest error on more than a half of the datasets. From the perspective of average rank, the proposed MA gains the rank $1.54 \pm 1.88$, which is the best one among the six methods investigated. Finally, the Wilcoxon signed rank test is used to verify the performance of the proposed approach against "CV" for each dataset[2]. The results show that the proposed MA is significantly superior to "CV" in terms of error rate on eight datasets. To be concluded, those results imply our proposed approach is competitive with other methods and even more effective and robust than others, which indicate that our proposed PSO-PS based MA is a practical method for parameters optimization in SVMs.

Table 2 Comparison of the variants of the proposed MA with PSO and GS

|  | PSO | | MA1 | | MA2 | | MA3 | | MA4 | | GS | |
| --- | --- | --- | --- | --- | --- | --- | --- | --- | --- | --- | --- | --- |
|  | error | #eva | error | #eva | error | #eva | error | #eva | error | #eva | error | #eva |
| *Banana* | 11.50±0.75 | **921.5**±87 | 11.03±0.68 | 997.1±**65** | 11.05±0.68 | 1031.2±73.6 | 10.75±0.52 | 1108±70.3 | **10.63±0.48** | 1029.7±75.5 | 11.9 | 1681 |
| *Breast cancer* | 26.73±5.16 | **987**±118.5 | 25.58±4.87 | 1058.6±**101** | 25.65±4.80 | 1135.8±119.4 | 24.91±**4.72** | 1165.8±126.7 | **24.69**±4.75 | 1109±115.3 | 25.97 | 1681 |
| *Diabetis* | 25.31±2.35 | **993.1**±87 | 24.41±2.30 | 1088.3±96 | 24.59±2.21 | 1126.6±105 | **23.47±1.98** | 1140.5±91.8 | **23.47±1.98** | 1110.8±94.4 | 23.67 | 1681 |
| *Flare solar* | 33.50±2.27 | **1026.5**±94.5 | 33.31±2.06 | 1071.5±87.7 | 33.18±2.20 | 1093.7±109.5 | 32.71±2.01 | 1184.3±103.3 | **32.56±1.8** | 1055.2±**80.6** | 34.25 | 1681 |
| *German* | 28.03±2.96 | **1061**±103.5 | 25.66±2.50 | 1153.6±93.5 | 25.49±2.73 | 1128.6±114.4 | 24.36±2.43 | 1109.5±98.1 | **24.01±2.26** | 1079.7±**91.5** | 24.33 | 1681 |
| *Heart* | 17.34±3.9 | **901.5**±108 | 16.85±3.26 | 985.3±99.7 | 16.74±3.26 | 1047.6±**89.6** | 16.18±2.87 | 994.4±93.6 | **16.13±2.45** | 976.8±90.7 | 19 | 1681 |
| **Ave** | 23.74±2.90 | **981.8**±99.8 | 22.81 ±2.61 | 1059.1±**90.5** | 22.78 ±2.65 | 1093.9±101.9 | 22.06 ±2.42 | 1117.1±93.6 | **21.92 ±2.29** | 1060.2±90.7 | 23.19 | 1681 |
| **AveRank** | 5.00±5.00 | **1**±3.5 | 3.83±3.50 | 2.8±2.3 | 3.83±3.50 | 4.2±3.8 | 2.08±1.75 | 4.5±3.2 | **1.08±1.25** | 2.5±2.17 | 4.33 | 6 |

Table 3 Average computational time in seconds of each method

|  | PSO | MA1 | MA2 | MA3 | MA4 | GS |
| --- | --- | --- | --- | --- | --- | --- |
| *Banana* | **126.35** | 129.89 | 147.81 | 145.76 | 149.57 | 187.8 |
| *Breast cancer* | **56.71** | 59.69 | 62.48 | 62.01 | 67.25 | 74.5 |
| *Diabetis* | 275.03 | 288.73 | **264.02** | 278.56 | 311.61 | 358.13 |
| *Flare solar* | 405.88 | **402.13** | 447.31 | 474.37 | 432.12 | 564.97 |
| *German* | **687.53** | 700.78 | 764.22 | 697.11 | 787.2 | 994.01 |
| *Heart* | **34.54** | 35.94 | 40.34 | 35.21 | 37.07 | 53.64 |
| **Ave** | **264.34** | 269.53 | 287.70 | 282.17 | 297.47 | 372.18 |
| **AveRank** | **1.33** | 2.50 | 3.67 | 3.00 | 4.50 | 6.00 |

## 5. Conclusions

The parameters of SVMs are of great importance to the success of the SVMs. To optimize the parameters properly, this study proposed a PSO and pattern search (PS) based Memetic Algorithm. In the proposed PSO-PS based MA, PSO was used to explore the search space and detect the potential regions with optimum solutions, while pattern search (PS) was used to conduct an effective refinement on the potential regions obtained by PSO. To balance the exploration of PSO and exploitation of PS, a probabilistic selection strategy was also introduced. The performance of proposed PSO-PS based MA was confirmed through experiments on several benchmark datasets. The results can be summarized as: (1) with different kinds of selection strategy, the variants of the proposed MA are all superior to single PSO and GS in terms of error rates and standard deviations, which confirms the success of refinement with pattern search. (2) Although the number of fitness evaluations and computational time of each MA are slightly inferior to PSO, each MA consumes much less fitness evaluations and computational time than grid search. (3) Among the variants of the proposed MA, the MA with our proposed probabilistic selection strategy gains the best performance in terms of error rates and standard deviations. (4) Compared with established counterparts in previous studies, the proposed PSO-PS based MA using proposed probabilistic selection strategy is capable of yielding higher and more robust performance than others. Hence, the proposed PSO-PS based MA can be a promising alternative for SVMs parameters optimization.

---

[2] Only "CV" and "Proposed MA" are compared using Wilcoxon signed rank test, because the detail results of others approaches on each dataset are not available. The detail results of "CV" on each dataset are available at http://www.raetschlab.org/Members/raetsch/benchmark/ .

Table 4 Comparison with established algorithms

| Dataset | CV [40] | RM [13] | FE [13] | L1-CB [12] | L2-CB [12] | Proposed MA |
|---|---|---|---|---|---|---|
| *Banana* | 11.53±0.66 | 10.48±0.40 | 10.68±0.50 | 11.65±5.90 | 10.44±0.46 | **10.35±0.40**\*\*\* |
| *Breast Cancer* | 26.04±4.74 | 27.83±4.62 | 24.97±4.62 | 28.75±4.61 | 26.03±**4.60** | **24.37**±4.61\* |
| *Diabetis* | 25.53±1.73 | 34.56±2.17 | **23.16**±1.65 | 25.36±2.95 | 23.50±1.66 | 23.17±**1.35**\*\*\* |
| *Flare Solar* | 32.43±1.82 | 35.51±**1.65** | 32.37±1.78 | 33.15±1.83 | 33.33±1.79 | **32.07**±1.73\* |
| *German* | 23.61±2.07 | 28.84±**2.03** | 23.41±2.10 | 29.26±2.88 | 24.41±2.13 | 23.72±2.05 |
| *Heart* | **15.95**±3.26 | 21.84±3.70 | 16.62±3.20 | 16.94±3.71 | 16.02±3.26 | **15.95**±**2.11** |
| *Image* | 2.96±0.60 | 4.19±0.70 | 5.55±0.60 | 3.34±0.71 | 2.97±**0.45** | **2.90**±0.65 |
| *Ringnorm* | 1.66±0.12 | **1.65**±**0.11** | 2.03±0.25 | 1.68±0.13 | 1.68±0.12 | 1.66±0.12 |
| *Splice* | 10.88±0.66 | 10.94±0.70 | 11.11±0.66 | 10.99±0.74 | **10.84**±0.74 | 10.89±**0.63** |
| *Thyroid* | 4.80±2.19 | 4.01±2.18 | 4.4±2.40 | 4.55±2.10 | 4.20±**2.03** | **3.4**±2.07\*\*\* |
| *Titanic* | 22.42±**1.02** | 23.04±1.18 | 22.87±1.12 | 23.51±2.92 | 22.89±1.15 | **21.58**±1.06\*\*\* |
| *Twonorm* | 2.96±0.23 | 3.07±0.246 | 2.69±0.19 | 2.90±0.27 | 2.64±0.20 | **2.49**±**0.13**\*\* |
| *Waveform* | 9.88±0.43 | 11.10±0.50 | 10±0.39 | 9.78±0.48 | 9.58±**0.37** | **9.30**±**0.37**\*\*\* |
| **Ave** | 14.67±1.50 | 16.70±1.55 | 14.60±1.50 | 15.53±2.25 | 14.50±1.46 | **13.99**±**1.33** |
| **AveRank** | 3.38±3.73 | 4.62±3.69 | 3.62±3.42 | 4.81±5.31 | 3.04±2.96 | **1.54**±**1.88** |

\* Statistically significant at the level of 0.05 when compared to CV.
\*\* Statistically significant at the level of 0.01 when compared to CV.
\*\*\* Statistically significant at the level of 0.001 when compared to CV.
Bold values are the best ones for each dataset

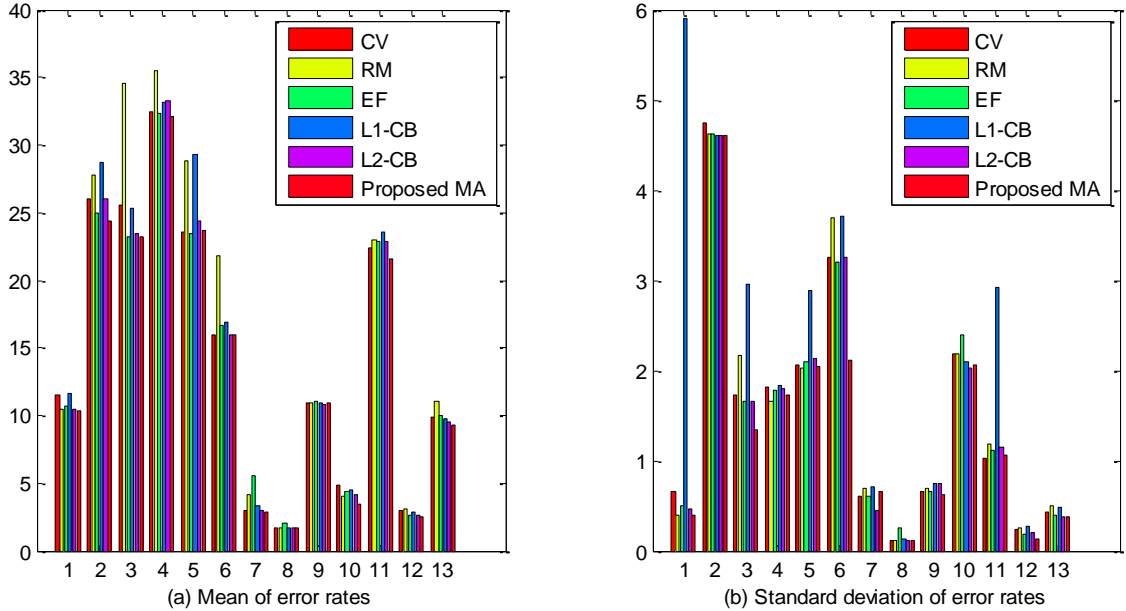

Fig.4. Results on 13 benchmark datasets: (a) Mean of error rates; (b) Standard deviation of error rates

One should note, however, that this study only focus on the PSO-PS based MA for parameters optimization in SVMs, some other varieties of PSO and evolutionary algorithms can also be used in the proposed MA. Besides, as the rapid development of hardware, the parallelization of the proposed MA to make full use of the increasingly vast computing resources (e.g., multiple cores, GPU, cloud computing) is another interesting topic. Other topics include investigating adaptive parameters setting of the proposed algorithms, comparing more extensively with other existing EAs or MAs, and developing more efficient memetic algorithms in dealing with other designing issues, e.g., population management, intensity of refinement [29]. Future work will be on the research of the above cases and applications of the proposed approach in practical problems.

## Acknowledgement

The authors would like to thank the anonymous reviewers for their valuable suggestions and constructive comments. This work was supported by the Natural Science Foundation of China (Grant No. 70771042) and the Fundamental Research Funds for the Central Universities (2012QN208-HUST) and a grant from the Modern Information Management Research Center at Huazhong University of Science and Technology.

## References

[1] V.N. Vapnik, The Nature of Statistical Learning Theory, Springer, New York, 1995.
[2] C.W. Hsu, C.C. Chang, C.J. Lin, A practical guide to support vector classification, Department of Computer Science, National Taiwan University, 2003.
[3] G. Moore, C. Bergeron, K.P. Bennett, Model selection for primal SVM, Machine Learning, 85 (2011) 175-208.
[4] O. Chapelle, V. Vapnik, O. Bousqet, S. Mukherjee, Choosing Multiple Parameters for Support Vector Machines, Machine Learning, 46 (1) (2002) 131-159.
[5] S.S. Keerthi, Efficient tuning of SVM hyperparameters using radius/margin bound and iterative algorithms, IEEE Transactions on Neural Networks, 13 (5) (2002) 1225-1229.
[6] K. Duan, S.S. Keerthi, A.N. Poo, Evaluation of simple performance measures for tuning SVM hyperparameters, Neurocomputing, 51 (2003) 41-59.
[7] K.M. Chung, W.C. Kao, C.L. Sun, L.L. Wang, C.J. Lin, Radius margin bounds for support vector

machines with the RBF kernel, Neural Computation, 15 (2003) 2643-2681.
[8] N.E. Ayat, M. Cheriet, C.Y. Suen, Automatic model selection for the optimization of SVM kernels, Pattern Recognition, 38 (10) (2005) 1733-1745.
[9] M.M. Adankon, M. Cheriet, Optimizing resources in model selection for support vector machine, Pattern Recognition, 40 (3) (2007) 953-963.
[10] L. Wang, P. Xue, K.L. Chan, Two Criteria for Model Selection in Multiclass Support Vector Machines, IEEE Transactions on Systems Man and Cybernetics Part B-Cybernetics, 38 (6) (2008) 1432-1448.
[11] M.M. Adankon, M. Cheriet, Model selection for the LS-SVM. Application to handwriting recognition, Pattern Recognition, 42 (12) (2009) 3264-3270.
[12] S. Li, M. Tan, Tuning SVM parameters by using a hybrid CLPSO–BFGS algorithm, Neurocomputing, 73 (2010) 2089-2096.
[13] Z.B. Xu, M.W. Dai, D.Y. Meng, Fast and Efficient Strategies for Model Selection of Gaussian Support Vector Machine, IEEE Transactions on Systems Man and Cybernetics Part B-Cybernetics, 39 (5) (2009) 1292-1307.
[14] F. Imbault, K. Lebart, A stochastic optimization approach for parameter tuning of Support Vector Machines, in: J. Kittler, M. Petrou, M. Nixon (Eds.), Proceedings of the 17th International Conference on Pattern Recognition, Vol 4, IEEE Computer Soc, Los Alamitos, 2004, pp.597-600.
[15] X.C. Guo, J.H. Yang, C.G. Wu, C.Y. Wang, Y.C. Liang, A novel LS-SVMs hyper-parameter selection based on particle swarm optimization, Neurocomputing, 71 (2008) 3211-3215.
[16] H.J. Escalante, M. Montes, L.E. Sucar, Particle Swarm Model Selection, Journal of Machine Learning Research, 10 (Feb) (2009) 405-440.
[17] X. Zhang, X. Chen, Z. He, An ACO-based algorithm for parameter optimization of support vector machines, Expert Systems with Applications, 37 (9) (2010) 6618-6628.
[18] T. Gomes, R. Prudencio, C. Soares, A. Rossi, A. Carvalho, Combining meta-learning and search techniques to select parameters for support vector machines, Neurocomputing, 75 (2012) 3-13.
[19] M.N. Kapp, R. Sabourin, P. Maupin, A dynamic model selection strategy for support vector machine classifiers, Applied Soft Computing, 12 (8) (2012) 2550-2565.
[20] P. Moscato, M.G. Norman, A `memetic' approach for the traveling salesman problem implementation of a computational ecology for combinatorial optimization on message-passing systems, Parallel Computing and Transputer Applications, (1992) 177-186.
[21] P. Moscato, On Evolution Search Optimization Genetic Algorithms and Martial Arts: Towards Memetic Algorithms, Caltech Concurrent Computation Program, C3P Report, 826 (1989).
[22] J. Tang, M.H. Lim, Y.S. Ong, Diversity-adaptive parallel memetic algorithm for solving large scale combinatorial optimization problems, Soft Computing, 11 (9) (2007) 873-888.
[23] S. Areibi, Z. Yang, Effective Memetic Algorithms for VLSI design = genetic algorithms + local search plus multi-level clustering, Evolutionary Computation, 12 (3) (2004) 327-353.
[24] A. Elhossini, S. Areibi, R. Dony, Strength Pareto Particle Swarm Optimization and Hybrid EA-PSO for Multi-Objective Optimization, Evolutionary Computation, 18 (1) (2010) 127-156.
[25] A. Lara, G. Sanchez, C. Coello, O. Schuetze, HCS: A New Local Search Strategy for Memetic Multiobjective Evolutionary Algorithms, IEEE Transactions on Evolutionary Computation, 14 (1) (2010) 112-132.
[26] Q.H. Nguyen, Y.S. Ong, M.H. Lim, A Probabilistic Memetic Framework, IEEE Transactions on Evolutionary Computation, 13 (3) (2009) 604-623.
[27] V.N. Vapnik, Statistical Learning Theory, John Wiley&Sons, Inc, New York, 1998.
[28] F.E.H. Tay, L.J. Cao, Application of support vector machines in financial time series forecasting, Omega, 29 (4) (2001) 309-317.
[29] X. Chen, Y. Ong, M. Lim, K. Tan, A Multi-Facet Survey on Memetic Computation, IEEE Transactions On Evolutionary Computation, 15 (5) (2011) 591-607.
[30] F. Neri, C. Cotta, Memetic algorithms and memetic computing optimization: A literature review, Swarm and Evolutionary Computation, 2 (2012) 1-14.
[31] X. Hu, R.C. Eberhart, Y. Shi, Engineering optimization with particle swarm, Proceedings of the 2003 IEEE Swarm Intelligence Symposium, 2003, pp.53-57.
[32] M. Reyes, C. Coello, Multi-objective particle swarm optimizers: A survey of the state-of-the-art, International Journal of Computational Intelligence Research, 2 (3) (2006) 287-308.
[33] J. Kennedy, R.C. Eberhart, Y. Shi, Swarm intelligence, Morgan Kaufmann Publishers, San Francisco, 2001.
[34] C. Chang, C. Lin, LIBSVM: a Library for Support Vector Machines, 2001. Software available at http://www.csie.ntu.edu.tw/~cjlin/libsvm.
[35] J. Kennedy, R.C. Eberhart, Particle Swarm Optimization, Proceedings of the IEEE International

Conference on Neural Networks, 4 (1995) 1942-1948.
[36] E.D. Dolan, R.M. Lewis, V. Torczon, On the local convergence of pattern search, SIAM Journal of Optimization, 14 (2) (2003) 567-583.
[37] B. Liu, L. Wang, Y.H. Jin, An effective PSO-based memetic algorithm for flow shop scheduling, IEEE Transactions on Systems Man and Cybernetics Part B-Cybernetics, 37 (1) (2007) 18-27.
[38] M. Lozano, F. Herrera, N. Krasnogor, D. Molina, Real-coded memetic algorithms with crossover hill-climbing, Evolutionary Computation, 12 (3) (2004) 273-302.
[39] M.L. Tang, X. Yao, A memetic algorithm for VLSI floorplanning, IEEE Transactions on Systems Man and Cybernetics Part B-Cybernetics, 37 (1) (2007) 62-69.
[40] G. Rätsch, T. Onoda, K.R. Müller, Soft Margins for AdaBoost, Machine Learning, 42 (3) (2001) 287-320.

Conference on Neural Networks, 4 (1995) 1942-1948.
[36] E.D. Dolan, R.M. Lewis, V. Torczon, On the local convergence of pattern search, SIAM Journal of Optimization, 14 (2) (2003) 567-583.
[37] B. Liu, L. Wang, Y.H. Jin, An effective PSO-based memetic algorithm for flow shop scheduling, IEEE Transactions on Systems Man and Cybernetics Part B-Cybernetics, 37 (1) (2007) 18-27.
[38] M. Lozano, F. Herrera, N. Krasnogor, D. Molina, Real-coded memetic algorithms with crossover hill-climbing, Evolutionary Computation, 12 (3) (2004) 273-302.
[39] M.L. Tang, X. Yao, A memetic algorithm for VLSI floorplanning, IEEE Transactions on Systems Man and Cybernetics Part B-Cybernetics, 37 (1) (2007) 62-69.
[40] G. Rätsch, T. Onoda, K.R. Müller, Soft Margins for AdaBoost, Machine Learning, 42 (3) (2001) 287-320.